\documentclass[runningheads]{llncs}
\usepackage{graphicx}
\usepackage{float}
\begin{document}
\title{Using Graph Neural Networks to Reconstruct Ancient Documents}
%
%
\author{Cecilia Ostertag\inst{1,2} \and
Marie Beurton-Aimar\inst{2}}
\authorrunning{C. Ostertag and M. Beurton-Aimar}
%
\institute{L3i - EA 2118, LaRochelle, France \email{cecilia.ostertag1@univ-lr.fr} \and
LaBRI - CNRS 5800 Bordeaux, France \email{beurton@labri.fr}}
\maketitle              
\begin{abstract}
In recent years, machine learning and deep learning approaches such as artificial neural networks have gained in popularity for the resolution of automatic puzzle resolution problems. Indeed, these methods are able to extract high-level representations from images, and then can be trained to separate matching image pieces from non-matching ones. These applications have many similarities to the problem of ancient document reconstruction from partially recovered fragments. In this work we present a solution based on a Graph Neural Network, using pairwise patch information to assign labels to edges representing the spatial relationships between pairs. This network classifies the relationship between a source and a target patch as being one of Up, Down, Left, Right or None. By doing so for all edges, our model outputs a new graph representing a reconstruction proposal. Finally, we show that our model is not only able to provide correct classifications at the edge-level, but also to generate partial or full reconstruction graphs from a set of patches.

\keywords{deep learning \and graph neural networks  \and document reconstruction \and cultural heritage}
\end{abstract}
\section{Introduction}

    The study of ancient documents and artifacts provides invaluable knowledge about previous civilizations. Indeed, they contain information about the economic, religious, and political organization at their time of writing. Unfortunately, the preservation conditions of such artifacts were often less than ideal, and nowadays archaeologists can only recover partial fragments of papers, papyri, and pottery shards. The restoration of old documents from recovered fragments is a daunting task, but it is necessary in order to decipher the writings. Recent image processing techniques can be applied to this problem, for example for text extraction \cite{wadhwani2020text} or object reconstruction \cite{rasheed2015survey}. In particular Convolutional Neural Networks, which are the best suited deep learning architectures for image inputs, can be used as an enhancement of traditional methods.
    
    Here we consider that the task of ancient document reconstruction is analogous to puzzle resolution \cite{kleber2009survey}, with the added challenge that the shape of the pieces cannot be used to help the reconstruction, as the fragments' edges were eroded and distorted with time. We describe image reconstruction from patches as an edge classification problem, where all patches are represented by nodes in a graph. Instead of using a classical approach of pairwise comparisons followed by global reconstruction with a greedy algorithm, we use a Convolutional Graph Neural Network to test every pair of patches in a single pass. We also provide an interactive user interface for manual checking and correction of the assembly graphs predicted by our model. 

\section{Related Works}

    In 2010, Cho et. al. \cite{cho2010probabilistic} introduce a graphical model to solve the problem of puzzle reconstruction. Their approach uses the sum-of-squared color difference across junctions as a compatibility metric to evaluate pairwise patch assemblies, and a Markov network to specify constraints for the whole-image reconstruction. An other pairwise compatibility metric that was used in similar applications is the Mahalanobis Gradient Compatibility \cite{mondal2013robust}. The work of \cite{jin2014jigsaw} use a more complex compatibility metric, taking into account edge similarity and cosine similarity between patches' histograms. In 2015, Paikin et. al. \cite{paikin2015solving} present an optimized greedy algorithm for solving jigsaw puzzles with patches of unknown orientation and with missing pieces, where the key point is to find the best starting patch for the global assembly. 
    
    All of these works deal with puzzles representing natural images, which carry a lot of semantic information (foreground, background, objects, people,...) and a diversity of shapes and colors. In our case of ancient documents reconstruction, the images contain less information ; mostly a low gradient of color across the whole image, some writings or inscriptions, and the texture of the document itself. Indeed, in the reconstruction examples given in all of these works, the areas that are the least correctly assembled are areas containing few semantic information, like the sky, the sea, or grass fields. A work by Paumard et. al. \cite{paumard2018jigsaw} seemed closer to our cultural heritage problem, since the authors' goal is to reconstruct old paintings, or pictures of ancient objects, from patches, but their dataset is also mainly made of images with important semantic information.
    
    In our previous work \cite{ostertag2020matching}, based on \cite{paumard2018jigsaw} approach using a Siamese Convolutional Neural Network to test pairwise assemblies, we proposed a model able to predict for a pair of patches the direction of assembly (up, down, left, or right), or its absence. Then we greedily constructed a graph by choosing good candidates for each pair, according to a probability threshold. In case of multiple candidates, several graphs were created, to give the user multiple assembly proposals. The major drawback of our approach was the execution time of our greedy algorithm, and the exploding number of assembly proposals generated by our reconstruction pipeline. 
    
    In this work we apply the increasingly popular Graph Neural Network architecture to our old document reconstruction problem, in order to test in a single shot the entirety of pairwise alignments between patches. Graph Neural Networks \cite{wu2020comprehensive} are used to work on non-Euclidean data. Here our original data is Euclidean, but we transform the set of image patches into a complete graph, hence the use of a Graph Network. In the context of classification, these networks can be used in different ways: node classification, edge classification, or graph classification, but always rely on node attributes to compute features. According to \cite{wu2020comprehensive}, Graph Neural Networks have mainly been applied to citation networks, bio-chemical graphs, and social networks, but to the best of our knowledge they haven't yet been used in the context of puzzle solving or document reconstruction.  

\section{Automatic Reconstruction using a ConvGNN}

    \subsection{Creation of a Ground Truth Dataset}
    
    Our image data consists in 5959 papyrus images, taken from the University of Michigan Library's papyrology collection, containing documents dating from 1000 BCE to 1000 CE \cite{michigan}. This dataset was also used by \cite{pirrone2019papy} with the aim of finding fragments belonging to the same original papyrus. From this dataset, we removed 1865 images corresponding to reverse sides of some papyri. These images don't contain any textual information, so are of no importance for egyptologists, and would risk to worsen the training of our model because of this lack of information.
    
    \begin{figure}[H]
    \includegraphics[width=\textwidth]{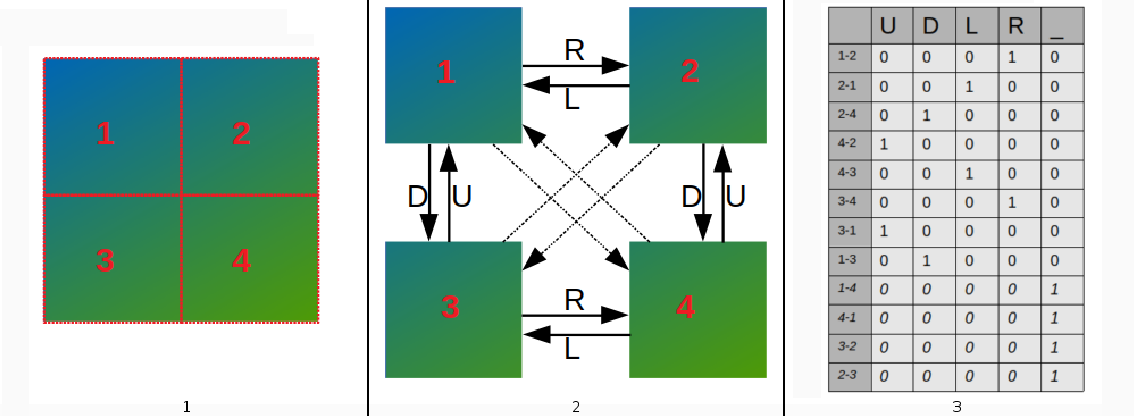}
    \caption{Example of ground truth creation from an image split into 4 patches. 1: Original image. 2: Complete directed graph, Nodes are patches and Edges are spatial relationships. 3: One-hot labels for all edges. U=up, D=down, L=left, R=right, \_=no relationship} \label{gt}
    \end{figure}
    
    We created our own dataset with ground truth by artificially splitting images from papyrus images into patches. Each image was first re-sized to 768$\times$1280 pixels, then split into 15 non-overlapping patches of size 256$\times$256. Then a graph representation of the image was created: this representation is a complete directed graph, where nodes are patches and edges represent the spatial relationship between two nodes. As per standard graph vocabulary, for a pair of nodes $(s, t)$ linked by an edge $e$ starting at node $s$ and ending at node $t$, we call $s$ the source node and $t$ the target node. Each node has as attribute an identifier and the pixel array corresponding to the patch image. Each edge has as attribute an identifier corresponding to the spatial relationship: up, down, left, right, and none. The edge attributes correspond to the five classes in our task, and can be encoded as one-hot vectors (see Fig.~\ref{gt}). 
    
    Given the nature of complete graphs, the class imbalance towards edges belonging to class ``none'' (see Table~\ref{samples}) would only increase with an increase of the number of patches per image. This is an important fact that we had to take into account during training.
   
   \begin{table}[H]
   \centering
       \caption{Number of samples per class, per image and for the entire dataset}
    \begin{tabular}{|l|c|c|c|c|c|}
    \hline
    \textbf{Class}                                                       & U     & D     & L     & R     & \_     \\ \hline
    \textbf{\begin{tabular}[c]{@{}l@{}}Samples\\ per image\end{tabular}} & 12    & 12    & 10    & 10    & 166    \\ \hline
    \textbf{\begin{tabular}[c]{@{}l@{}}Samples \\ total\end{tabular}}    & 49128 & 49128 & 40940 & 40940 & 679604 \\ \hline
    \end{tabular}
    \label{samples}
    \end{table}

    \subsection{Model Architecture}
    
\begin{figure}[H]
    \includegraphics[width=\textwidth]{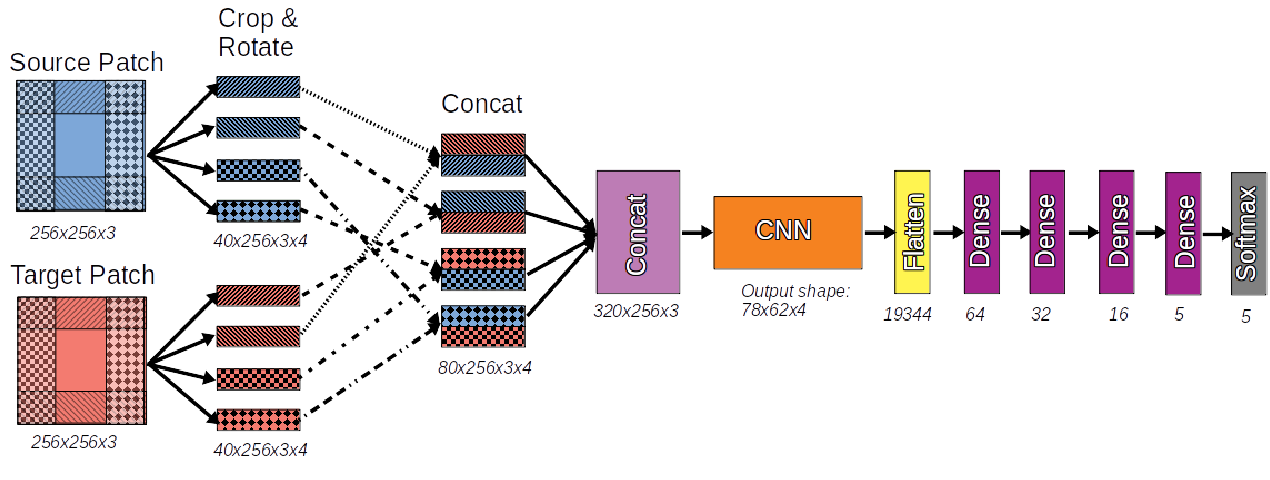}
    \caption{Architecture of the Pairwise Comparison Model part of our AssemblyGraphNet} \label{PairwiseComparisonModel}
    \end{figure}

    Our model, that we named AssemblyGraphNet, is a Convolutional Graph Neural Network that uses node attributes information to predict edge labels. In order to do this, the model is actually made of two parts: a Global Model, and a Pairwise Comparison Model. The Pairwise Comparison Model (see Fig.~\ref{PairwiseComparisonModel}) is a convolutional neural network which takes as input two connected nodes (source and target), and outputs a label for the edge connecting the source to its target. Here we defined five classes, corresponding to the alignment direction of the target patch in relation to the source patch. The classes are: up, down, left, right, and none.
    
    As we focus on testing pairwise alignments in 4-connectivity, the first step of the network is to crop a 40-pixel-width stripe in each direction, for both patches. Then these stripes are concatenated according to the possible matching sites (for example, target-up with source-down). The four ``assembly sites" are then concatenated together, and fed to a convolutional neural network (see Table~\ref{layers}) followed by four Dense layers of decreasing size. Finally, a Softmax layer outputs the probability that the edge corresponds to each of the five classes described previously. We note that we created our ``assembly sites" using a concatenation operation, and not an averaging or an absolute difference, in order to benefit from the asymmetry inherent to this operation. Indeed, our goal is to check pairwise assembly in both directions, to obtain more information for the global reconstruction step. 
    
    \begin{table}[H]
    \centering
    \caption{Architecture and parameters of the convolutional neural network used in the Pairwise Comparison Model}
    \begin{tabular}{|l|c|c|c|c|}
    \hline
    \multicolumn{1}{|c|}{\textbf{Layer}} &\textbf{Output Shape} & \textbf{Kernel Size} & \textbf{Stride} & \textbf{Output Nodes} \\ \hline
    BatchNorm                            & 320 $\times$ 256 $\times$  3 &                     &                 &                      \\ \hline
    Convolution                          & 318 $\times$  254 $\times$  4  & 3                   & 1               & 4                    \\ \hline
    ReLU                                 & 318 $\times$  254 $\times$  4 &                     &                 &                      \\ \hline
    MaxPooling                           & 159 $\times$  127 $\times$  4 & 2                   & 2               &                      \\ \hline
    BatchNorm                            & 159 $\times$  127 $\times$  4 &                     &                 &                      \\ \hline
    Convolution                          & 157 $\times$ 125 $\times$  4 & 3                   & 1               & 4                    \\ \hline
    ReLU                                 & 157 $\times$  125 $\times$  4  &                     &                 &                      \\ \hline
    MaxPooling                           & 78 $\times$  62 $\times$  4  & 2                   & 2               &                      \\ \hline
    BatchNorm                            & 78 $\times$  62 $\times$  4 &                     &                 &                      \\ \hline
    \end{tabular}
    \label{layers}
    \end{table} 
    
    The Global Model is the part that actually uses the graph representation of the image. The graph connectivity is represented by a matrix of size [2 $\times$ number of edges], where the first row is the identifiers of the source nodes and the second row is the identifiers of the target nodes. The RGB pixel values for all patches are stored in a node features matrix, and the edge labels are all initialized to [1,1,1,1,1] and stored into an edge features matrix. Using this connectivity matrix, two graph-level feature matrix of shape [number of edges $\times$  patch size $\times$  patch size $\times$  3] are inferred from the node features matrix: one for the source nodes, and the other for the target nodes. The Pairwise Comparison Model is used on every source-target pair, while handling the first dimension of the matrix as if it were the dimension of a batch. The output of the Global Model is a new graph where the connectivity matrix and the node features matrix are not changed, but the edge feature matrix is replaced by the predicted edge labels (argmax of the Pairwise Comparison Model's output for each edge).

    \subsection{Experiments and Results}
    
    Our model was implemented using pytorch and torch-geometric \cite{Fey/Lenssen/2019}, a pytorch library for graph neural networks. For memory and computation efficiency, the computations were made using float16 precision. The loss function is the categorical cross-entropy computed between the ground truth and predicted edge feature matrix, and the optimizer is Adam. We split our dataset into 3394 training images, 500 validation images, and 200 test images. During training the order of node pairs is shuffled for each image and at each iteration, to avoid an overfitting caused by the order of appearance of edges.
    
    As we explained earlier, we have a huge class imbalance towards the fifth class, so to remedy this problem we decided to weight the loss for each class. A weight of 0.1 was assigned to the fifth class, and a weight of 0.8 for each of the remaining classes.  During training and validation, we computed the balanced accuracy (average of recall obtained on each class) instead of the classic accuracy (percentage of correct predictions), to take into account class imbalance and have a score more representative of our model's ability to predict correct pairwise assemblies \cite{brodersen2010balanced,kelleher2020fundamentals}. We also computed the F1-score for each class, as 1-versus-all binary classification results.

    \begin{figure}[H]
    \includegraphics[width=\textwidth]{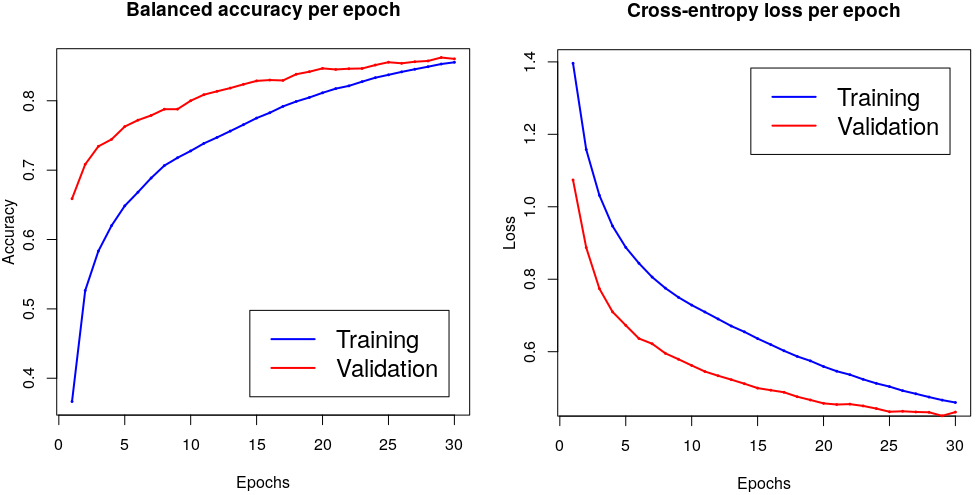}
    \caption{Evolution of balanced accuracy and categorical cross-entropy loss during training} \label{loss}
    \end{figure}

    Our results (see Fig.~\ref{loss} and Table~\ref{training}) obtained after 30 epochs show that we reach a limit with a satisfying value of 86\% balanced accuracy on the validation set. However, even though we used a weighted loss, the fifth class, corresponding to ``no patch assembly" is the most correctly predicted. We can safely assume that a part of the features extracted by the Pairwise Comparison Model are related to the gradient of pixel values at the junction zones. This explains why the fifth class remains the best predicted class, given that it is easy to differentiate between a high gradient (corresponding to an absence of assembly) and a low gradient (corresponding to a possible assembly), but difficult to infer the direction of the match from such information.  
    
    \begin{table}[H]
    \centering
    \caption{Value of loss, balanced accuracy, and per-class F1-score obtained after training for 30 epochs}
    \label{training}
    \begin{tabular}{|l|c|c|c|c|c|c|c|}
    \hline
    \multicolumn{1}{|c|}{\textbf{}} & \textbf{Loss} & \textbf{Acc} & \textbf{F1\_1} & \textbf{F1\_2} & \textbf{F1\_3} & \textbf{F1\_4} & \textbf{F1\_5} \\ \hline
    \textbf{Training}               &    0.46     &   0.86      &     0.65       &    0.63       &     0.50      &     0.51      &     0.81       \\ \hline
    \textbf{Validation}             &    0.43      &    0.86     &    0.69        &      0.67      &    0.54       &      0.52      &    0.83       \\ \hline
    \textbf{Test}             &     0.42     &  0.85       &    0.69       &     0.68      &    0.53       &    0.52       &    0.81       \\ \hline
    \end{tabular}
    \end{table}
    
    Even though our results are convincing, they are based on the pairwise classification of patch pairs, and not on the actual whole-image reconstruction. The problem for evaluating the quality of image reconstructions is that with our model each node is susceptible to have multiple neighbour candidates in each direction. One solution is to filter automatically the results based on a threshold on the class-associated probabilities, and to consider every alignment probability inferior to this threshold as an occurrence of the ``no patch assembly" class, but there is a risk to delete alignments correctly predicted but without enough confidence. Instead of this, we opted for an interactive visualization of the assembly proposal graphs, allowing the user to delete some ``obviously wrong to the human eye" edges.

\section{Interactive Visualization of Assembly Proposals}

    \subsection{Graphical Interface}
    
    After running our AssemblyGraphNet on a set of patches, the resulting graph is saved in a text file, with all node features, and predicted edge labels with associated probabilities. To provide the user with a lightweight graphical interface, we use the Cytoscape.js graph library \cite{franz2016cytoscape} to read and display the graph elements in an interactive way. To see correctly the reconstructed image, each node is attributed a 2D set of coordinates, and represented on the screen by the associated pixel values of the corresponding patch. 
    
    First the connected components are identified, then for each component, a node is arbitrarily chosen as the origin, and the coordinates of the rest of the nodes are iteratively computed using the spatial relationships to one-another, following a depth first search algorithm. We note that this can lead to several nodes being in the same position, and that the user needs to be careful to check if this happens. The connected components are separated from each other, to show clearly the partial image reconstructions.
    
    The user can then choose between a compact representation, with edges and labels hidden, or an expanded representation showing node identifiers as well as edges with their label and associated probabilities. The user is able to move nodes around freely, to manually set a threshold for edge filtering, to select edges for deletion, and finally to refresh the node positions after modifications in the graph. 
    
    \subsection{Examples of Image Reconstructions}
    
\begin{figure}[H]
    \includegraphics[width=\textwidth]{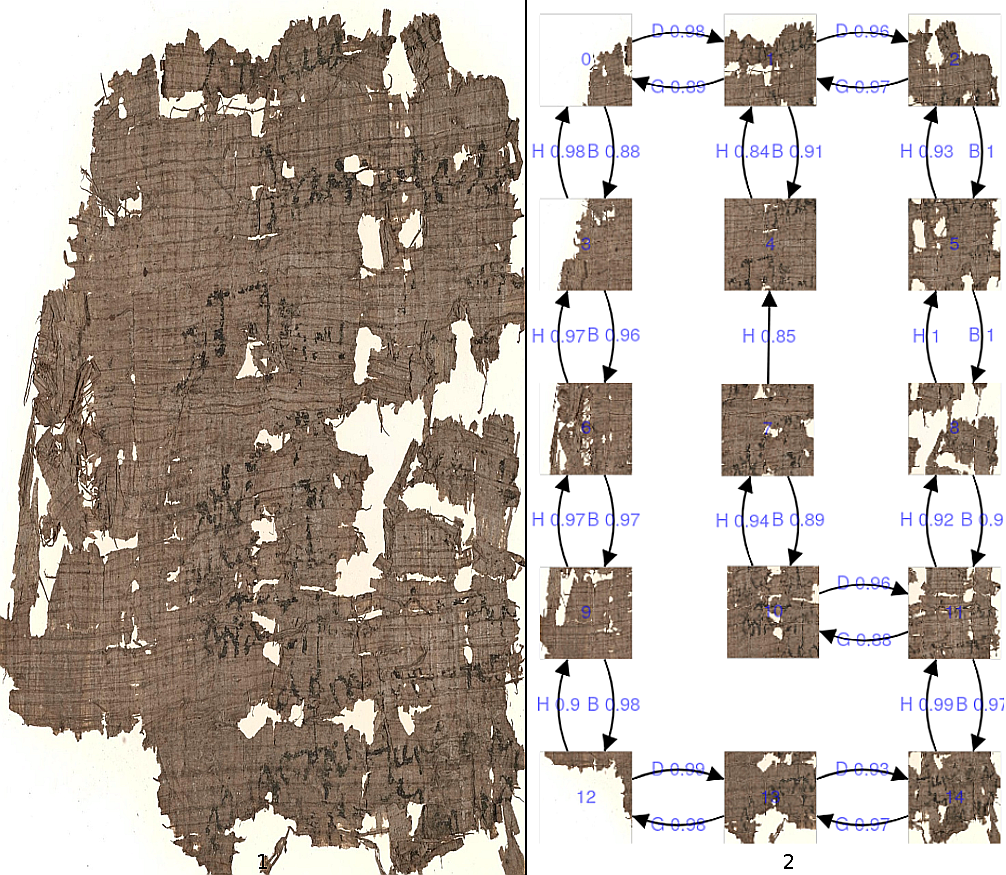}
    \caption{Example of entirely correct image reconstruction. 1: Ground truth. 2: Assembly graph.} \label{bien}
    \end{figure}

\begin{figure}[H]
    \centering
    \includegraphics[scale=0.42]{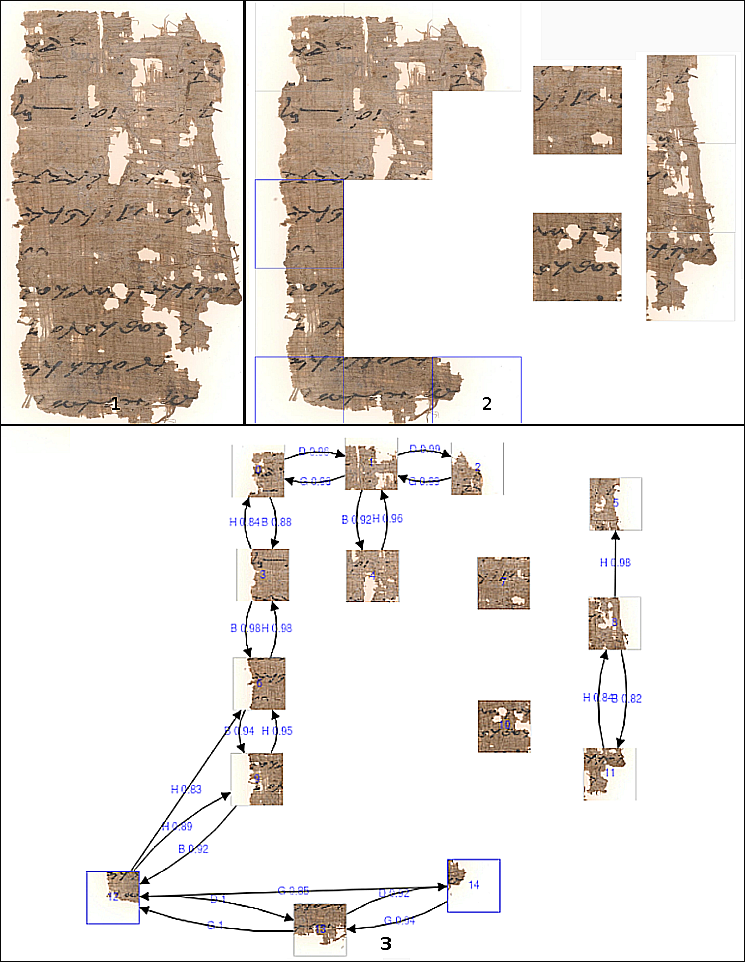}
    \caption{Example of partial image reconstruction. 1: Ground truth. 2: Reconstructions. 3: Assembly graph.} \label{bof}
    \end{figure}

    Using images from our test set, we used our visualization tool to assess the quality of our assembly proposals. We started by inputting lists of 15 patches from the same image, and looking at reconstruction results for individual images. Fig.~\ref{bien} and Fig.~\ref{bof} show assembly proposals, after filtering edges with a probability threshold of 0.8 for all classes. The assembly graph in Fig.~\ref{bien} shows a case where we obtain a perfect reconstruction even if all edges were not correctly predicted, or not predicted with a satisfying probability. It also demonstrate the use of using a directed graph to test pairwise alignments both ways, as it means that we have twice the chance to have a correct prediction for each pair.  
    
    Fig.~\ref{bof} illustrates cases where only partial assemblies were found, either because of edge labels incorrectly predicted as ``no patch assembly", or because of an edge filtering that was too drastic. In this assembly graph we have two nodes, highlighted by a blue border, that each have multiple candidates for the same alignment direction: patch 6 and 9 are both candidates to be above patch 12, and patch 12 and 13 are both candidates to be left to patch 14. However, it is easy for the user to filter out the incorrect edges, based simultaneously on the prediction probabilities, and on the existence of the reciprocal edge label in the target to source direction. Here for example the edge 12 $\rightarrow$ 6 has a probability of 0.83, inferior to the edge 12 $\rightarrow$ 9 with a probability of 0.89, but the relationship between patches 12 and 6 could also be removed on the basis that edge 6 $\rightarrow$ 12 with label ``up" doesn't exist but edge 9 $\rightarrow$ 12  with label ``up" does. The same reasoning can be applied to filter out edge 12 removed on the basis that edge 6 $\rightarrow$ 12 with label ``up" doesn't exist but edge 9 $\rightarrow$ 14, and to finally obtain a better reconstruction.  
    
    \subsection{Reconstructions from multiple images}

    \begin{figure}[H]
    \includegraphics[width=\textwidth, height=\textheight]{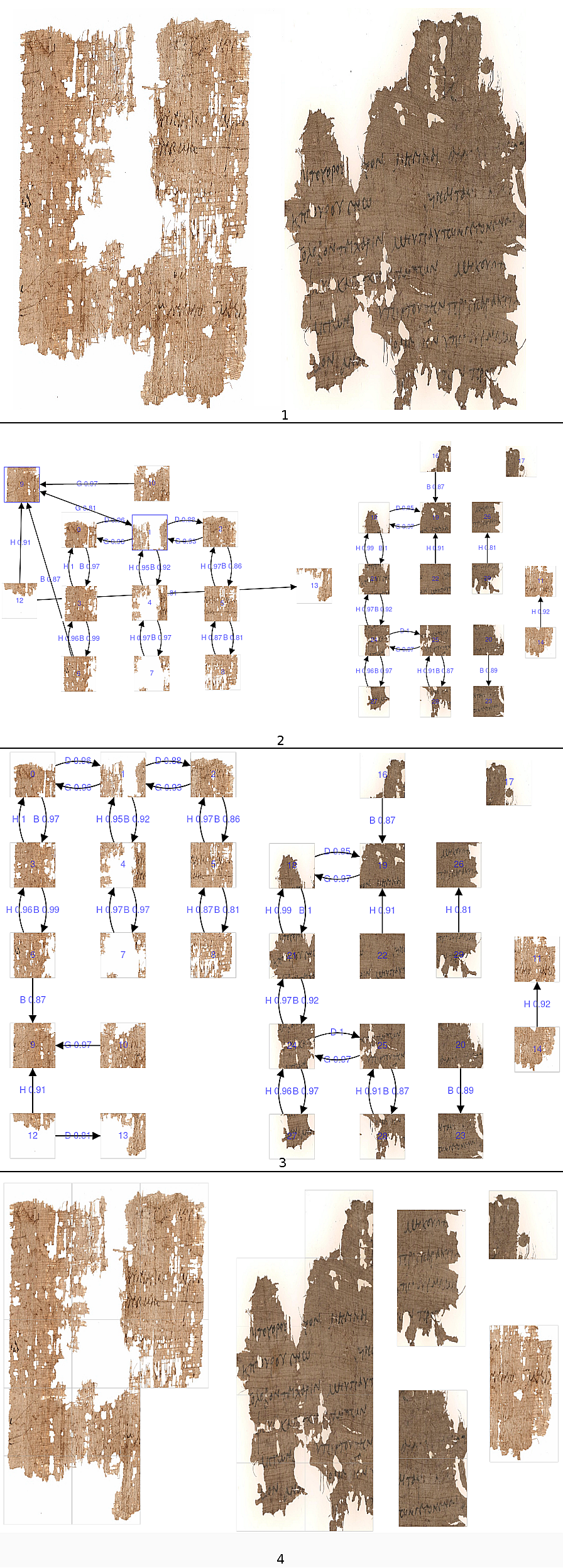}
    \caption{Example of partial reconstructions from patches belonging to two images. 1: Ground truth. 2: Assembly graph before user intervention. 3: Assembly graph after user intervention (deletion of edge 4 $\rightarrow$ 9). 4: Reconstructions.} \label{mix1}
    \end{figure}
    
    After showing the ability of our model to provide accurate reconstructions of an image from the corresponding set of patches, we wanted to see if our model was capable to distinguish between patches coming from different images. To do this, we simply fed the network with a random list of patches coming from two test images.
    
        \begin{figure}[H]
    \includegraphics[width=\textwidth]{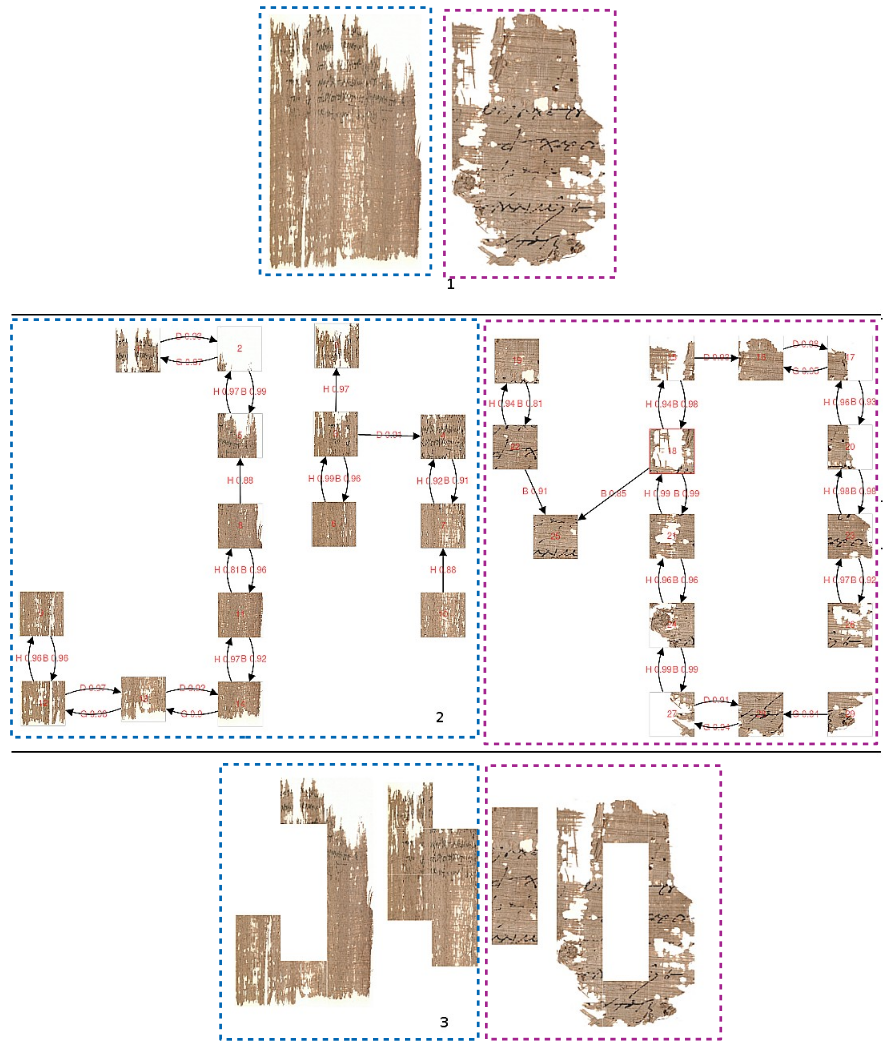}
    \caption{Example of partial reconstructions from patches belonging to two images. 1: Ground truth. 2: Assembly graph, 3: Reconstructions.} \label{mix2}
    \end{figure}

    Fig.~\ref{mix1} and Fig.~\ref{mix2} are two examples, after filtering the probabilities at 0.8. The first figure shows a case where the two images are easily distinguished by the color of the papyrus, and the second where the differences are more subtle, being papyrus texture and scribe handwriting. Fig.~\ref{mix1} also illustrate the importance of user input, as in this case a single edge was identified and removed by the user, instead of several edges including important ones if we had opted for automatic thresholding.
    
    Looking at the assembly graph for these two examples, we can see that, with a 0.8 probability threshold, patches from different images are not mixed together. Moreover, even if the images are not wholly reconstructed, the user is provided with partial assemblies.

\section{Discussion and Conclusion}

    Our model AssemblyGraphNet gives an average of 86\% correctly predicted spatial relationships between patch pairs for the validation set. This is a significant improvement over results obtained by Paumard et. al. \cite{paumard2018jigsaw}, who obtained 68.8\% of correctly placed fragments for their dataset of pictures from the Metropolitan Museum of Art, composed mainly of paintings. Moreover, the authors of this article say that they obtain a perfect reconstruction only 28.8\% of the time. We don't have an objective means of estimating the number of perfect reconstructions that we can obtain, given that we rely on user intervention to filter the assembly graphs, but with our edge classification score we can assume that a larger number of perfect reconstructions can be obtained by using our model. Finally, in a situation where patches come from multiple images, the model proposed by Paumard et. al. will not be able to separate the images during global reconstruction because they don't use a class to model the absence of assembly, contrarily to our AssemblyGraphNet.  

    By taking advantage of the Graph Neural Network architecture, our model is able to simultaneously predict the existence and, if need be, the direction of assembly between all pairs given a set of patches. Thus, the model outputs directly a whole-image reconstruction proposal. For the time being the model doesn’t make use of a lot of graph properties, with the exception of the connectivity matrix allowing to process all node pairs at once. However this graphNN-based approach was used with the intent of reconstructing the whole images in several steps instead of a single step like we presented here, using a succession of convolution layers followed by graph coarsening layers \cite{wu2020comprehensive} to assemble groups of small fragments into bigger fragments, that will be assembled together in the next layer, and so on.
    
    Then, thanks to our interactive visualization interface, a user with some knowledge of the field can refine the assembly graph based on the probabilities associated to edge labels, or based on symmetric and neighbouring edges. Moreover, if we consider only the inference, the execution time of our model is quicker than the execution of a greedy algorithm based on pairwise compatibility: on a GeForce RTX 2080 Ti, it takes an average of 0.8 ms to output an assembly graph from a set of 15 patches.   

    We also showed that our AssemblyGraphNet is not only able to predict spatial relationships between patches in an image, but is also able to distinguish between different images, based on information such as color, texture, and writings. This result is important because it means that in a real-life application there will be no need for a preliminary step to group together fragments belonging to the same initial document, which is a problem tackled by \cite{pirrone2019papy}.  
    
    At this time inputs of arbitrary shapes could be used with our model, with a padding to reconcile the input shape as a square or rectangle for the CNN architecture, but it would introduce noise. For example if we use zero-padding, then the gradient between two padded inputs would be small, even if the texture or color of the actual fragment were different. To tackle this problem, a strategy would be to introduce such input cases at training time. An other strategy would be to process the whole patches instead of the border regions only. Finally, the background (pixels corresponding to padded regions) versus foreground (pixels corresponding to the actual fragment) membership could be encoded into the network, so that during training the network learns to discard background-pixels information.
    
    Here we consider that the correct orientation of every patch is known, so we will not test the possible rotation of patches. In practice, this assumption could only be made for patches containing writings, using the disposition of the symbols to find the correct rotation beforehand. In future works, we plan to work on solutions to take into account the possible rotation of patches, as well as erosion at the border of the fragments. We note that in this work the extraction of border stripes from the patches was done because of memory constraints (as explained before), but actually using the whole patches as input to a Siamese CNN would provide more information, and importantly information not limited to the junction zones between fragments, as these junctions might not correspond to anything if there was a lot of erosion.

%
%
%
\bibliographystyle{splncs04}
\bibliography{biblio.bib}

\end{document}